%% file: main.tex
\documentclass[a4paper,fleqn]{cas-sc}

\usepackage{graphicx}
\usepackage{caption}
\usepackage{subcaption}
\usepackage[numbers]{natbib}
\def\tsc#1{\csdef{#1}{\textsc{\lowercase{#1}}\xspace}}
\tsc{WGM}
\tsc{QE}
\tsc{EP}
\tsc{PMS}
\tsc{BEC}
\tsc{DE}

\begin{document}
\let\WriteBookmarks\relax
\def\floatpagepagefraction{1}
\def\textpagefraction{.001}
\shorttitle{Support Vector Regression with Butterfly Optimization Algorithm}
\shortauthors{Mohammadreza Ghanbari, Hamidreza Arian}

\title [mode = title]{Forecasting Stock Market with Support Vector Regression and Butterfly Optimization Algorithm}                      

\author[1,3]{Mohammadreza Ghanbari}


\address[1]{Department of Mathematical Sciences , Sharif University of Technology, Tehran, Iran}

\author[2,4]{Hamidreza Arian}[style=chinese]

\cormark[1]
\ead{Hamidreza.Arian@uToronto.ca}

\ead[url]{www.hamidarian.com}

\address[2]{Graduate School of Management and Economics, Sharif University of Technology, Tehran, Iran}


\address[3]{RiskLab Khatam, Khatam University, Tehran, Iran}

\address[4]{RiskLab Toronto, Toronto, Canada}

\cortext[cor1]{Corresponding author}

\begin{abstract}
Support Vector Regression (SVR) has achieved high performance on forecasting future behavior of random systems. However, the performance of SVR models highly depends upon the appropriate choice of SVR parameters. In this study, a novel BOA-SVR model based on Butterfly Optimization Algorithm (BOA) is presented. The performance of the proposed model is compared with eleven other meta-heuristic algorithms on a number of stocks from NASDAQ. The results indicate that the presented model here is capable to optimize the SVR parameters very well and indeed is one of the best models judged by both prediction performance accuracy and time consumption.

\end{abstract}




\maketitle

\section{Introduction}

The problem of forecasting stock price movements, due to market's uncertainty from incoming news, non-linear financial instruments and behavioral and emotional biases is a challenging task facing academics and practitioners in the field; perhaps by far more complex than predicting the course of a comet by a physicist.  In the past, many models have been proposed to face this problem including Support vector regression (SVR) as the extended routine designed from 
Support Vector Machines (SVM). Originally introduced by Vapnik for classification problems, SVM was redesigned to solve regression problems in the SVR framework. Nevertheless, SVM can solve small-sample, non-linear and high dimension problems by using the structural risk minimization principle instead of the empirical risk principle, which could theoretically guarantee to achieve the global optimum~\cite{Cortes1995}.

Although SVR experimental results have shown great performance compared to other non-linear methods~\cite{ Thissen2003, Vapnik1995}, its performance mainly depends on the choice of parameters. Using wrong set of parameters for SVR creates considerably poor performance~\cite{Chapelle2002, Duan2003,Kwok2000,  Yeh2011}. The selection procedure mainly is based upon either trial and error or optimization techniques. In general, grid search~\cite{Hsu2003}, gradient descent~\cite{Keerthi2007} and metaheuristics algorithms~\cite{Blum2003, Talbi2009} are the three common optimization techniques to optimize the SVR parameters.

Metaheuristics algorithms have been introduced as problem-independent routines for finding an optimum solution and depict superior results when solving optimization problems like parameter calibration of complex models~\cite{Gogna2013}. 
In the last two decades many metaheuristics 
methods have been proposed for the SVR optimal parameter selection. For example, Genetic Algorithms (GA)~\cite{Huang2012, Jirong2010, Min2006, Wu2009}, Particle 
Swarm Optimization (PSO)~\cite{Xia2011, Wu2010},
Artificial Bee Colony (ABC)~\cite{Hong2011},
Grey Wolf Optimizer (GWO)~\cite{Mustaffa2015},
Firefly Algorithm (FA)~\cite{ Kavousi2014, Kazem2013},
Bat Algorithm (BA)~\cite{Tavakkoli2015},
Moth-Flame Optimization (MFO)~\cite{Li2016} and
Sine Cosine Algorithm (SCA)~\cite{Li2018}.
Recently, a novel population-based optimization algorithm designed by Arora and Singh~\cite{Arora2019} proposed mimicking the food foraging behavior of butterflies by searching the solution space in the optimization algorithm. In Butterfly Optimization Algorithm (BOA), information is propagated to
all other butterflies using fragrance and forms a general knowledge system with some
information loss. This algorithm utilizes a probability parameter to make a decision on the 
movement direction of butterflies either towards the best solution or a random search. 

In this paper, the objective is to propose a novel BOA-SVR model where BOA is employed to optimize the parameters of SVR. 
For comparison, eleven other alternatives including Salp Swarm Algorithm (SSA)~\cite{Mirjalili2017},
Biogeography-Based Optimization (BBO)~\cite{Simon2008}, 
Sine Cosine Algorithm (SCA)~\cite{Mirjalili2016}, 
Particle Swarm Optimization (PSO)~\cite{Kennedy2010}, 
Harmony Search Optimization (HSO)~\cite{Geem2001}, 
Genetic Algorithm (GA)~\cite{Goldberg2006}, 
Firefly Algorithm(FA)~\cite{Ynag2010}, 
Invasive Weed Optimization(IWO)~\cite{Mehrabian2006}, 
Teaching-Learning-Based Optimization(TLBO)~\cite{Rao2011}, 
Crow Search Algorithm(CSA)~\cite{Askarzadeh2016} 
 and Artificial Bee Colony Optimization (ABC)~\cite{Karaboga2007} are also extended to estimate the parameters of SVR.

The remainder of the paper is organized as follows: In Section \ref{method}, we discuss the proposed BOA-SVR model, the support vector regression and BOA algorithm. In Section \ref{experiment}, the established model is tested on several datasets and compared 
with other models and the experimental results are discussed. Conclusions and future research are provided in Section \ref{con}.





\section{Methodology}\label{method}
In this section, first we start by a brief overview of data pre-processing Phase Space Reconstruction routine. Then we describe the Support Vector Regression and the Butterfly Optimization Algorithm.

\subsection{Phase Space Reconstruction}
Numerous studies in recent years have confirmed the existence of
chaotic behavior in real life time series.
Phase Space Reconstruction is one of popular methods to uncover the hidden information embedded in the time series dynamics. This routine reconstructs an $m$-dimensional phase space, providing a simplified multidimensional representation of data.

Let $ \{x_i\}_{i=1}^n$,
represent an $n$ point time series.
Then, the reconstructed phase space can be expressed as a matrix,
\begin{equation}
X =
\begin{bmatrix}\label{xtakens}
x_1 & x_{1+\tau} & \cdots &  x_{1+(m-1)\tau}\\
x_2 & x_{2+\tau} & \cdots & x_{2+(m-1)\tau}\\
 \vdots & \vdots & \ddots & \vdots \\
x_{n-1-(m-1)\tau} & x_{n-1-(m-2)\tau} & \cdots & x_{n-1}
\end{bmatrix},
\end{equation}
where $m$ is called the embedding dimension of reconstructed phase
space and $\tau$ is the time delay constant. 

According to Takens’ theorem
\cite{Takens1981}, sufficient condition for the embedding
dimension is 
$m \geq 2d+1$, with $d$ being dimension of the time series data.
An efficient method of finding the minimal sufficient embedding
dimension is the false nearest neighbors (FNN) procedure,
proposed by Kennel et al. 
\cite{Kennel1992}. This method finds the nearest
neighbors of every point in a given dimension, and
then, checks to see if these points are still close neighbors
in one higher dimension. To estimate the delay parameter, we use the first minimum of the Mutual Information (MI) function \cite{Abarbanel1996}:
\begin{equation}
\text{MI}(\tau) = \sum_{j=1}^{n-\tau}p(x_j,x_{j+\tau})\text{log}_2\big(\frac{p(x_j,x_{j+\tau})}{p(x_j)p(x_{j+\tau})} \big)
\end{equation}
where
$p(x_j)$, $p(x_j,x_{j+\tau})$
 are marginal and and joint
probability densities, respectively. After finding the optimal embedding dimension $m$ and delay time 
$\tau$, the input data and output vector were re-designed by Eq. \ref{xtakens} and Eq. \ref{ytakens}.
\begin{equation}\label{ytakens}
Y =
\begin{bmatrix} 
Y_1\\
Y_2\\
\vdots\\
Y_n 
\end{bmatrix}
=
\begin{bmatrix}
x_{2+(m-1)\tau}\\
x_{3+(m-1)\tau}\\
\vdots\\
x_n
\end{bmatrix} 
\end{equation}

\subsection{Support Vector Regression}

Support vector machine (SVM) is a machine learning algorithm first introduced by Vapnik (1995) for classification problems. SVM was later promoted to support vector regression (SVR) by using a new type of loss function called $\epsilon$-insensitive loss function which is used to penalize
data as long as they are greater than
$\epsilon$~\cite{Vapnik1995}. SVR is a non-linear kernel-based regression method which provides the best regression hyperplane with smallest structural risk in high dimensional feature space \cite{Yeh2011}.

Assume we are given a training
dataset 
$\{(x_i,y_i)\}_i^n$, where 
$x_i \in \mathbb{R}^n$
is input data,
$y_i \in \mathbb{R}$ is the output value of the $i$-th data point in the dataset, $d$ is the dimension of samples and $n$ is the number of samples. The SVR function is formulated as:
\begin{align} \label{2.1}
y=f(x)=w^{T} \phi(x)+b,
\end{align}
where $\phi$ denotes the non-linear mapping from the input
space to the feature space, $w$ is a vector of weight
coefficients and $b$ is a bias constant. The $w$ and $b$ are
estimated by minimizing the following optimization problem:
\begin{equation}
\begin{split}
\text{Min} & \frac{1}{2} ||w||^2, \\
\text{s.t.: } &
\begin{cases}
y_i-w^T\phi(x_i)-b  & \leq \epsilon,\\ 
y_i-w^T\phi(x_i)-b  & \geq -\epsilon,
\end{cases}
\end{split}
\end{equation}

To cope with feasibility issues and to make the method more robust, we use slack variables $\xi, \xi^*$ to penalize deviations from the SVR band:
\begin{equation}
\begin{split}\label{2.3}
\text{Min} & \frac{1}{2} ||w||^2+ C\sum_{i=1}^n(\xi_i+\xi_i^*),\\
\text{s.t.: } &
\begin{cases}
y_i-w^T\phi(x_i)-b  \leq \epsilon + \xi_i,\\ 
y_i-w^T\phi(x_i)-b  \geq -\epsilon - \xi_i^*,\\
\xi_i,\xi_i^* \geq 0, i = 1,\cdots, n,
\end{cases}
\end{split}
\end{equation}
where $C$ is a constant known as the penalty factor, $\epsilon$ is the insensitive loss parameter and the slack variables $\xi_i, \xi_i^*$, measure the amount of difference between the estimated value and the target value beyond $\epsilon$.

After using Lagrangian multipliers and conditions for optimality, we
find a model solution in dual representation to solve Eq. \ref{2.1}, write the solution as
\cite{smola2003, Vapnik1995}:
\begin{align}
f(x)=\sum_{i=1}^{n}(\beta_i-\beta_i^*)K(x_i,x)+b,
\end{align}
where $\beta_i, \beta_i^*$ are nonzero Lagrangian multipliers and 
$K(x_i,x)$
is the kernel function. Here, we use Radial Basis Function (RBF) as Kernel:
\begin{align}
K(x_i,x_j) = \exp(-\gamma||x_i-x_j||^2),
\end{align}
where $\gamma$ is the RBF width parameter.

\subsection{Butterfly Optimization Algorithm (BOA)}

The Butterfly Optimization Algorithm is a novel metaheuristic optimization algorithm
inspired by the food foraging behavior of butterflies
\cite{Arora2019}.
In this algorithm, each butterfly generates fragrance with some intensity which is
correlated with its fitness function value. When a butterfly is able to sense fragrance
from any other butterfly, it will move toward it by
\begin{equation}\label{notrand}
    x_i = x_i +(r^2 \times g^* -x_i)\times \mathit{f}_i
\end{equation}
On the other hand, when a butterfly is not able to sense fragrance of the others, then it will move randomly by
\begin{equation}\label{rand}
    x_i = x_i +(r^2 \times x_j -x_k)\times f_i 
\end{equation}
where $x_i, x_j$ and $x_k$ indicate the positions of $i$-th, $j$-th and $k$-th butterflies, respectively. Here,
$g^*$ represents the current best solution, $r$ is a random number in
$[0, 1]$ and $f$ is the perceived magnitude of the fragrance given by 
\begin{equation} \label{eqn:fragintensity}
    \mathit{f} = c \mathit{I}^ a
\end{equation}
with $c$ as the sensory 
modality, ${I}$ as the stimulus intensity and $a$ as the power exponent dependent on modality. Equations \ref{notrand} and \ref{rand} refer to the global and local search phases, respectively. In BOA a random number $p$ is used as a switch probability between
global and local search states. 

\begin{figure}
	\centering
	    \includegraphics[scale=.25]{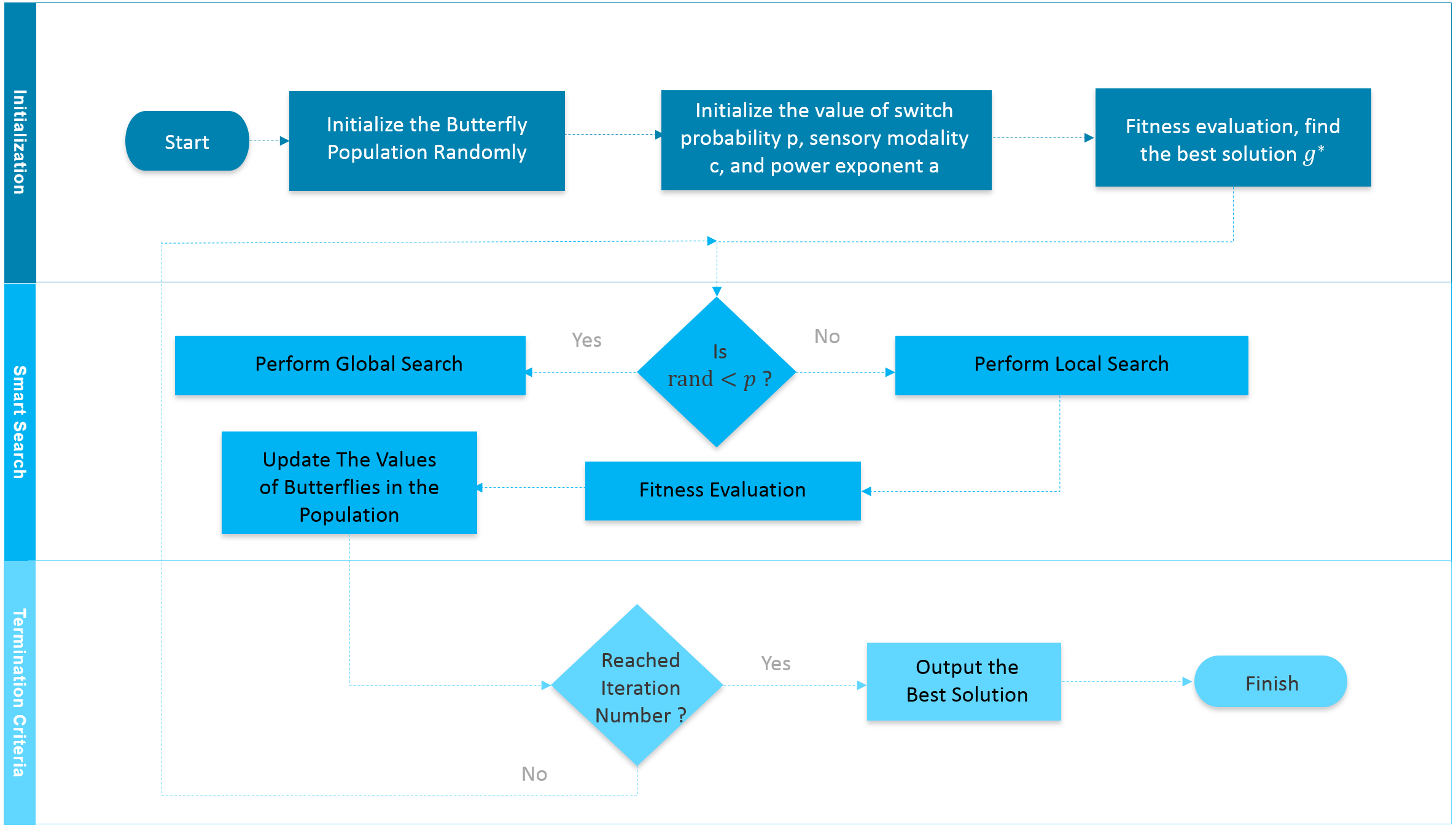}
	\caption{Flowchart of Butterfly Optimization Algorithm.}
	\label{figBAFlowChart}
\end{figure}

Fig. \ref{figBAFlowChart} presents the flow chart of the Butterfly Optimization algorithm. The entire procedure of BOA can be briefly explained in three phases: $(1)$ Initialization phase, $(2)$ Smart Search phase and $(3)$ Termination phase.
In the Initialization phase, positions of butterflies are generated
randomly and their fitness values are evaluated. In each iteration, in Smart Search phase, all butterflies move to new positions  by the Eq. \ref{notrand} and Eq. \ref{rand}, then their fitness values are re-evaluated. The
above procedure continues until the termination criteria is reached and the final best solution is given as output of the algorithm. 


\subsection{BOA for Parameter Calibration of SVR}

\begin{figure}
	\centering
	    \includegraphics[scale=.32]{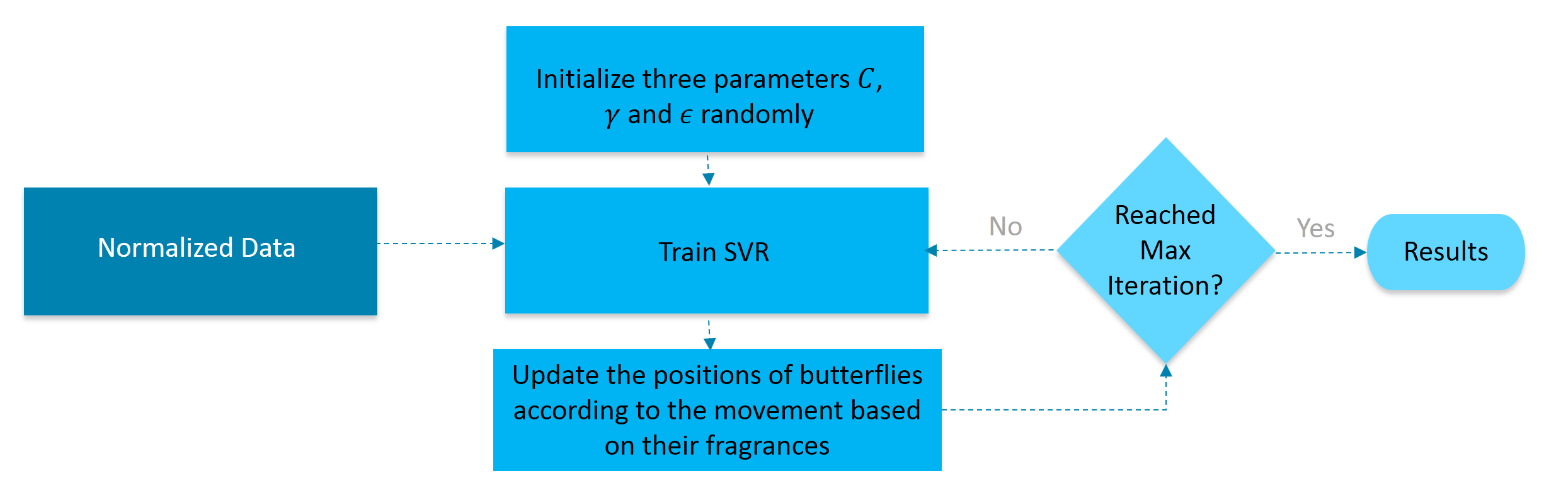}
	\caption{BOA-SVR procedure}
	\label{flochart}
\end{figure}

In the first step of our proposed model, the BOA parameters including the number of agents and maximum number of iterations are set. Then BOA-SVR starts with a set of candidate solutions generated randomly within predetermined lower and upper bounds. In this case, each solution is a three-dimensional vector represented by ($C$, $\gamma$, $\epsilon$) where   
$C$,
$\gamma$
and
$\epsilon$
are the SVR parameters to be optimized. The output of the tested SVR model is used as the fitness (objective) function. At the end,
if the number of iterations is equal to its maximum number, then the best solution is selected. Otherwise, the algorithm proceeds to the next iteration.
Fig. \ref{flochart}
shows the complete procedure of the proposed BOA-SVR.

The step-wise procedure of the proposed algorithm is described as follows:
\begin{itemize}[leftmargin=25mm]
\item[\textbf{Step 1:}]
Preprocess dataset with reconstructing time series phase space by estimation of time delay parameter
$\tau$ and optimum embedding dimension $m$. 
\item[\textbf{Step 2:}]
Normalize data using min-max formula
\begin{equation}\label{2.4}
x_{\text{new}} = \frac{x_{\text{old}}-x_{\text{min}}}{x_{\text{max}}-x_{\text{min}}}
\end{equation}
to scale to the range $[0,1]$. finally, divide the input data into the training set and
the testing set.
\item[\textbf{Step 3:}]
Assign the parameters including the number of search agents and the maximum number of iterations. Then set the iteration number, $t$, equal to zero.
\item[\textbf{Step 4:}]
Initialize the random solutions of search agents by  
\begin{equation}\label{2.5}
s_p = lb_p + (ub_p-lb_p)\cdot u
\end{equation}
and evaluate fitness function by using Eq. \ref{2.6} and Eq. \ref{2.7} on the test dataset. Here, $p \in \{C, \gamma, \epsilon\}$, and $lb$ $(ub)$ is lower (upper) bound, respectively and $u$ is a uniform random number in the interval $(0,1)$.
\item[\textbf{Step 5:}]
Calculate the fragrance of each butterfly. Update the positions of butterflies and the value of sensory modality, $a$, then evaluate the fitness function by training SVR. Set $t=t+1$.
\item[\textbf{Step 6:}]
If the maximum number of iterations is reached, output the optimal values of SVR parameters; otherwise go back to step 5.
\item[\textbf{Step 7:}]
Build the SVR model with the optimal parameters ($C$, $\gamma$, $\epsilon$) and make the 
prediction.
\end{itemize}

\section{Experimental results}\label{experiment}

In this section, a number of stocks are selected to evaluate the performance of our proposed BOA-SVR model. The proposed algorithm is compared to the other metaheuristic algorithms, available in the literature, which are utilized for parameter estimation of SVR, including
genetic algorithms based SVR (GA-SVR), 
particle swarm optimization based SVR (PSO-SVR), 
artificial bee colony based SVR (ABC-SVR), 
firefly algorithm based SVR (FA-SVR), 
salp swarm algorithm based SVR (SSA-SVR), 
harmony search optimization based SVR (HSO-SVR), 
invasive weed optimization based SVR (IWO-SVR), 
sine cosine algorithm based SVR (SCA-SVR), 
crow search algorithm based SVR (CSA-SVR), 
biogeography-based optimization based SVR (BBO-SVR) and teaching-learning based optimization algorithm based SVR (TLBO-SVR) . To the best of our judgement, this list of eleven metaheuristic algorithms consist the most popular metaheuristic routines available in the literature which are used for SVR parameter estimation. 

Facebook (from 4/25/2017 to 4/25/2019), Microsoft (from 4/24/2017 to 4/24/2019) and Intel (from 4/24/2017 to 5/8/2019) daily closing stock market prices were extracted from NASDAQ historical quotes available in the NASDAQ stock market. After finding time delay, $\tau$, embedding dimension, $m$ and reconstructing the phase space, the dataset was divided into two sets. $80\%$ of the
data were used as the training set and the remaining were used as the testing set. All the predictions were based on one-step ahead prediction results. The computations were carried out in MATLAB R2016a environment using the LIBSVM Toolbox \cite{Chang2011} on a laptop with an Intel(R) Core(TM) i3-3110M CPU @ 2.40GHz and 4 Gbytes memory.

In order to measure the prediction accuracy, we use MSE and MAPE measures.
\begin{align}
    MSE &= \frac{1}{N}\sum_{i=1}^{N}(y_i - f_i) \label{2.6}\\
    MAPE &= \frac{1}{N}\sum_{i=1}^{N}\big|\frac{y_i - f_i}{y_i} \big| \label{2.7}
\end{align}
where $y_i$ and $f_i$ denote the actual and predicted values for the $i$-th data point, respectively and $N$ is the number of forecasting days.

Since population-based optimization techniques search for the optimal value of the problem randomly, there is no guarantee of finding the optimal solution just by one single run. 
However, with a diverse population and a sufficiently large number of iterations, the probability of finding the global optimum increases. In this experiment,
the number of population is selected to be 20 and the maximum 
number of iterations is 50. Also the search space for both parameters $C$ and $\gamma$ is $[4^{-10}, 4^{4}]$ and the range for parameter $\epsilon$ is $[4^{-10}, 4^{-1} ]$.
The parameters of the BOA algorithm in the proposed model are experimentally set. Probability switch is 0.8, sensory modality is 0.01 and power exponent is 0.1.

Table
\ref{table 1}
presents the details of datasets including name, number of embedding dimension $m$ and 
number of time delay $\tau$ in each group after phase space reconstruction in which we used the recurrence plot and recurrence quantification analysis MATLAB toolbox
\cite{Chen2012}.

\input{tabs/phasespace.tex}

\begin{figure}
\centering
\hspace*{0.2in}
\begin{subfigure}{.4\textwidth}
  \includegraphics[width=1.3\linewidth]{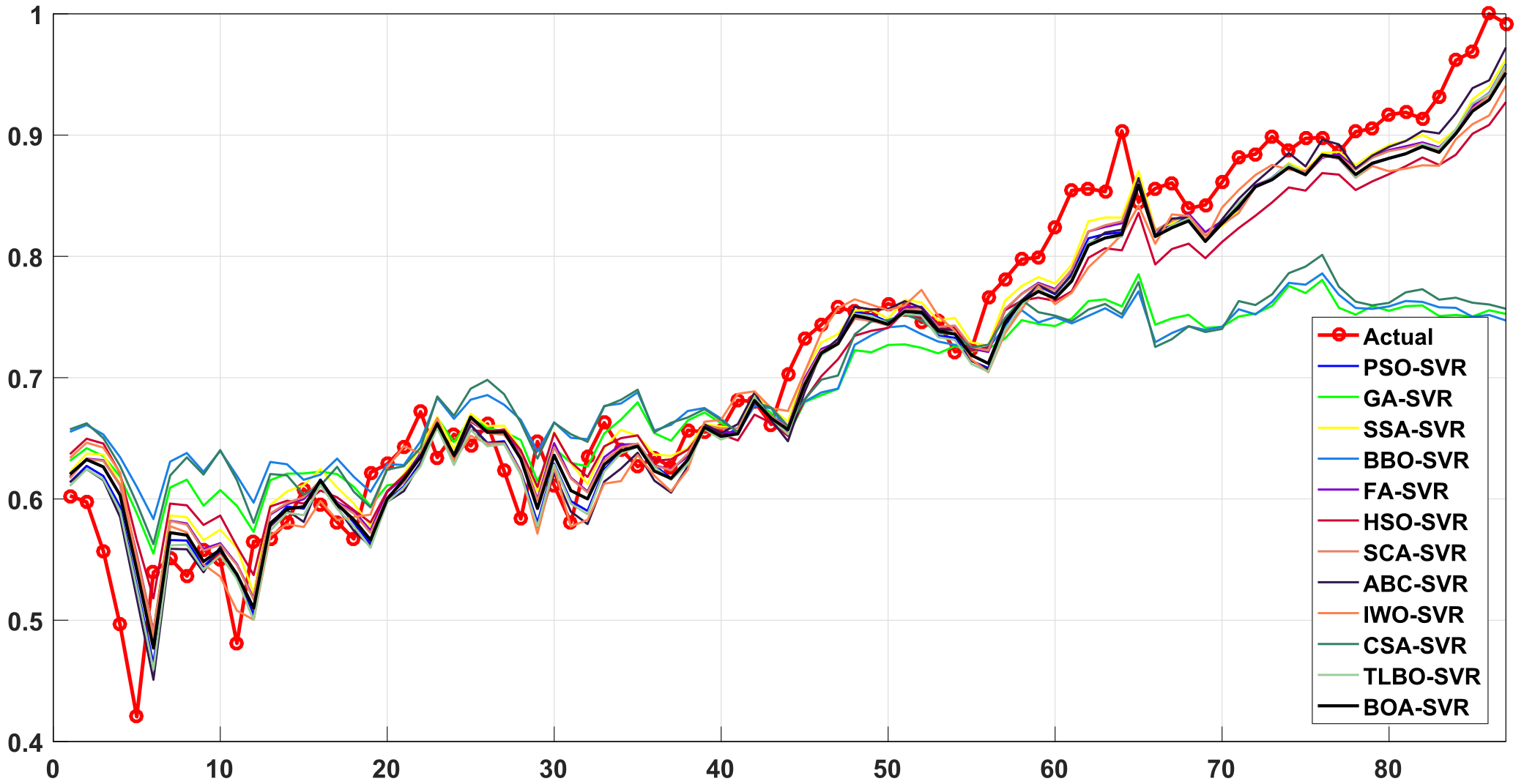}
  \caption{Microsoft}
  \label{microsofttests}
\end{subfigure}
\newline
\centering
\begin{subfigure}{.4\textwidth}
\hspace*{-0.9in}
  \includegraphics[width=1.3\linewidth]{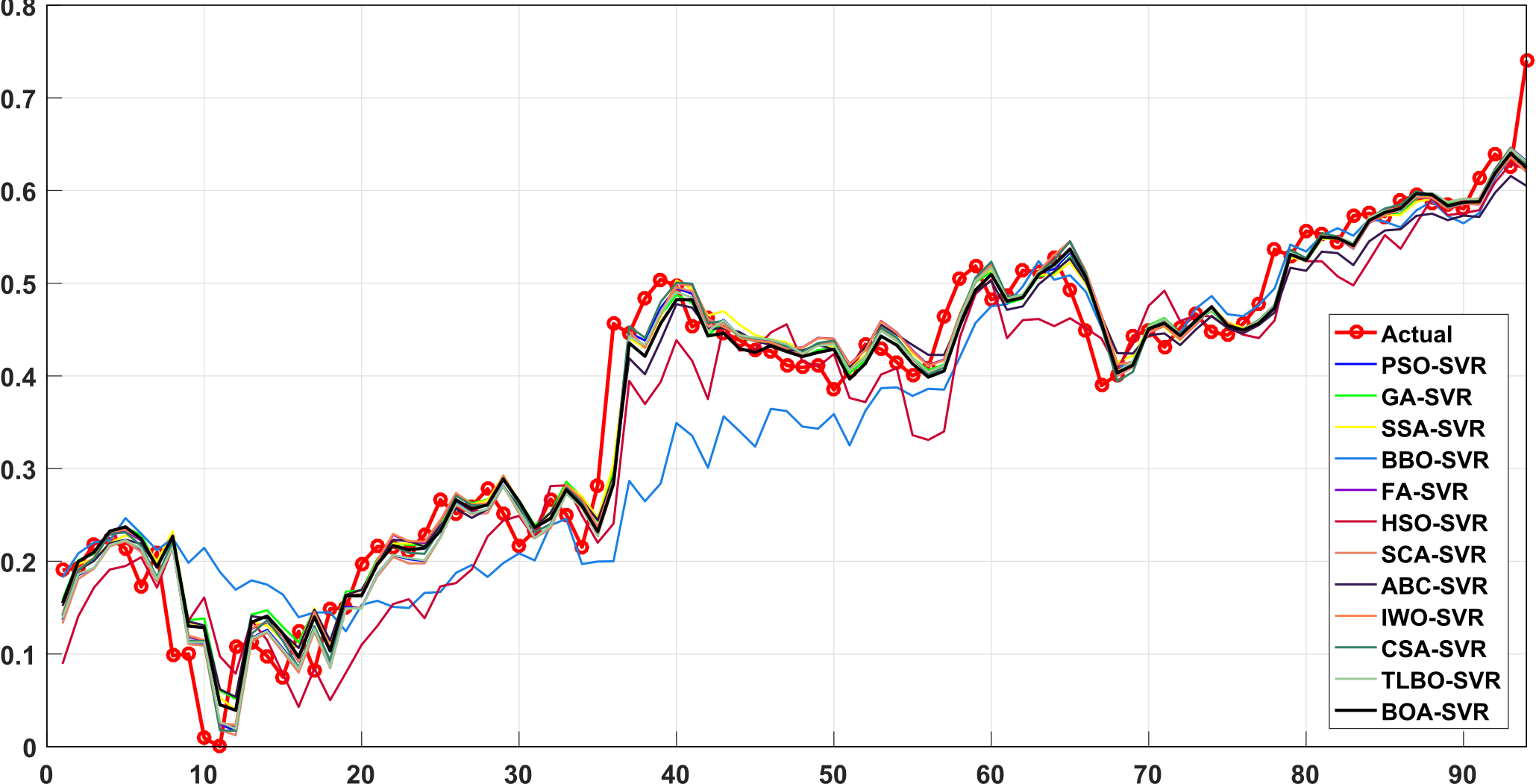}
  \centering \caption{Facebook}
    \label{facebooktests}
\end{subfigure}
\centering
\begin{subfigure}{.4\textwidth}
\hspace*{-0.1in}
  \includegraphics[width=1.3\linewidth]{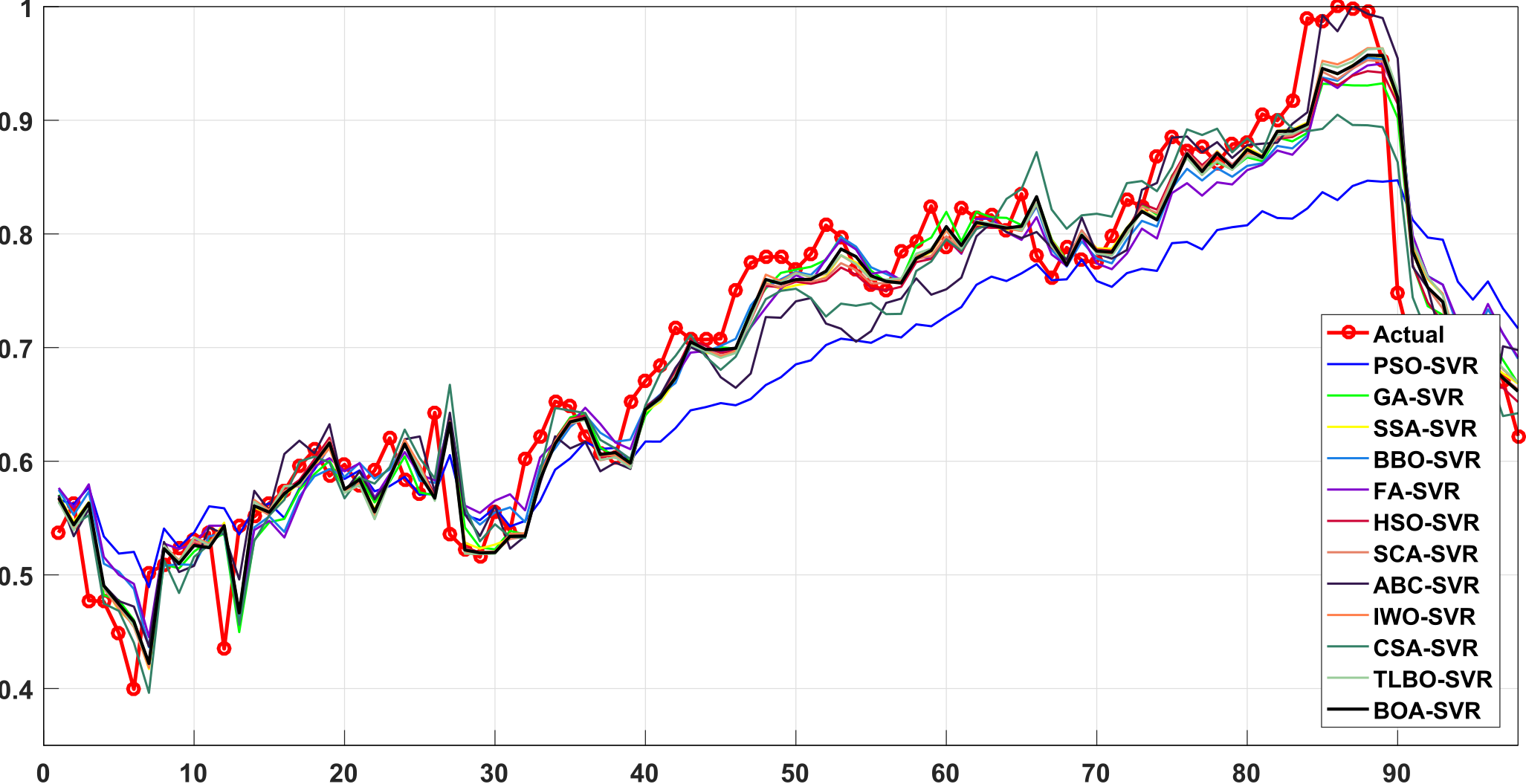}
  \centering \caption{Intel}
     \label{inteltests}
\end{subfigure}
\caption{Comparison of model forecasts for a) Microsoft b) Facebook and c) Intel show that among the twelve algorithms, BOA-SVR is consistently one of the best performers in terms of accuracy of prediction. }
\label{allTests}
\end{figure}

Fig. 
\ref{allTests}
illustrates the actual and predicted values of our model compared to eleven others for three testing stocks Microsoft (Fig. \ref{microsofttests}), Facebook (Fig. \ref{facebooktests}) and Intel (Fig. \ref{inteltests}). The corresponding parameters used for the above mentioned results for Microsoft is shown in Table
\ref{mictable}. This table presents the optimal value for the three parameters
$C, \gamma$ and $\epsilon$, as well as mean squared error (MSE), mean absolute
percentage error (MAPE) and the time consumption in testing sets for all of SVR-based methods. Tables \ref{facetable} and \ref{inteltable}, present the same results for Facebook and Intel, respectively.

\begin{figure}
\centering
\begin{subfigure}{.4\textwidth}
\hspace*{-0.9in}
  \includegraphics[width=1.3\linewidth, height=4.5cm]{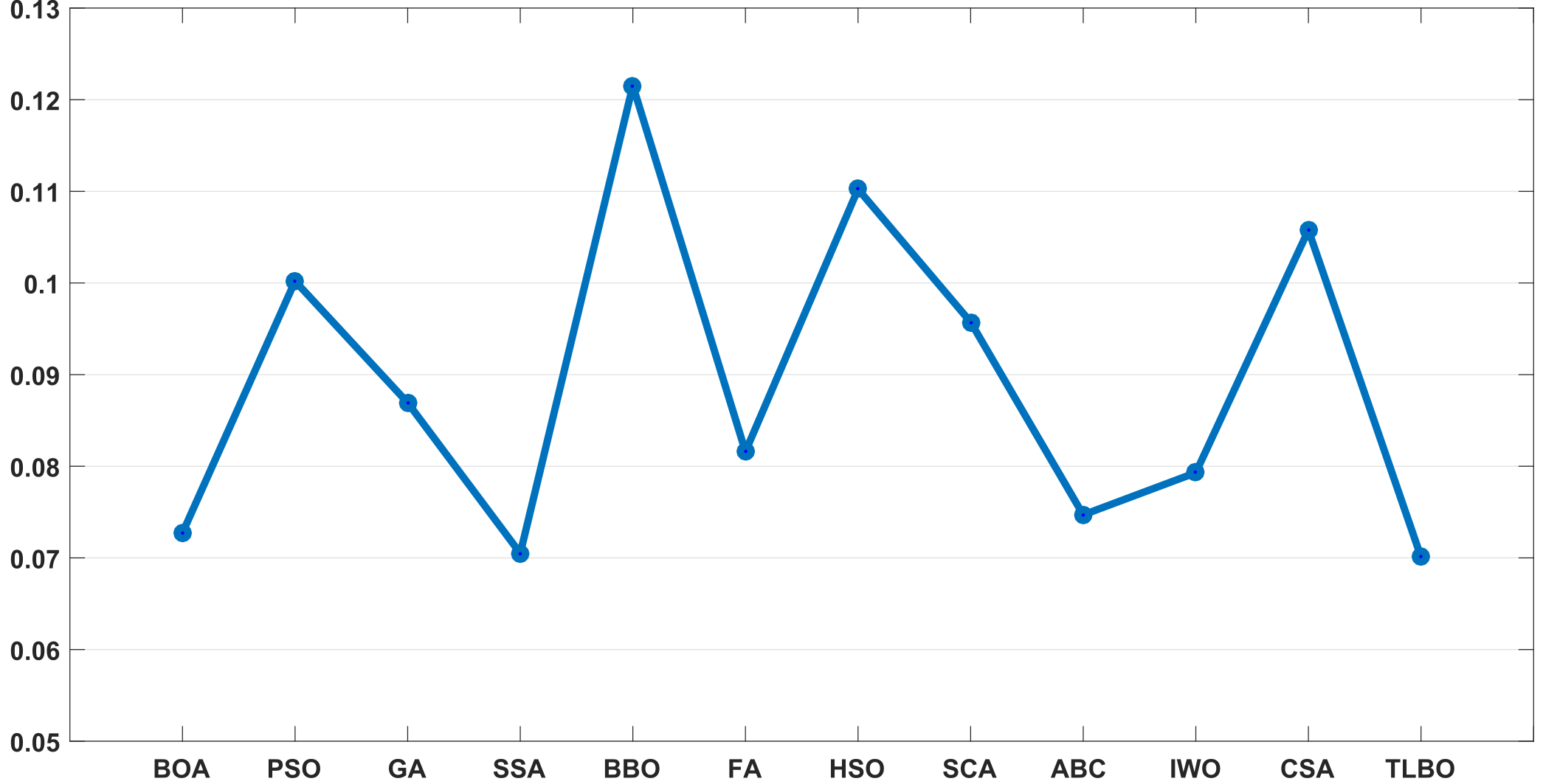}
  \centering \caption{MAPEs}
     \label{mape}
\end{subfigure}
\begin{subfigure}{.4\textwidth}
  \includegraphics[width=1.3\linewidth,height=4.5cm]{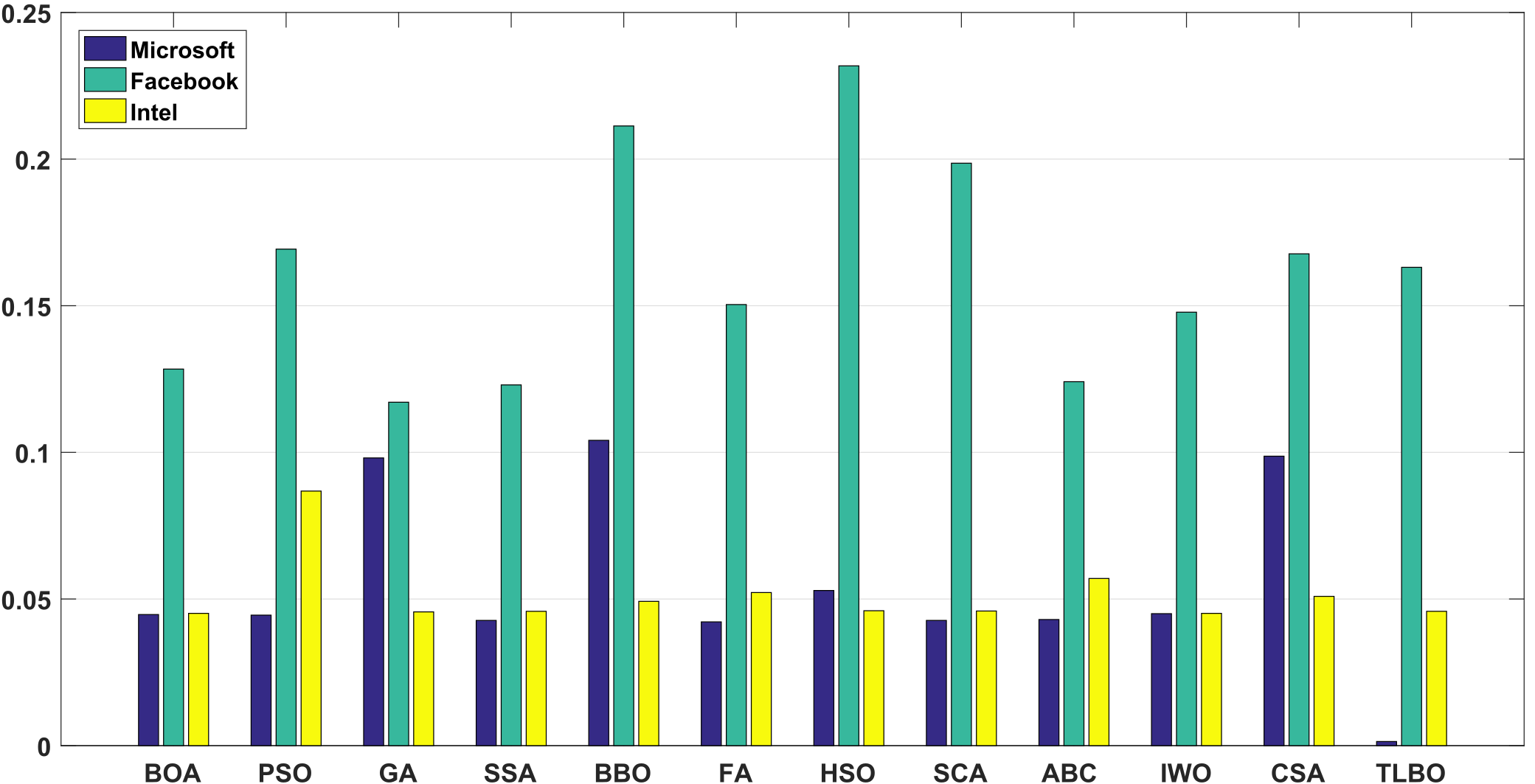}
  \centering \caption{Bar plot of MAPEs}
  \label{mapebar}
\end{subfigure}
\begin{subfigure}{.4\textwidth}
\centering
\hspace*{-0.9in}
  \includegraphics[width=1.3\linewidth, height=4.5cm]{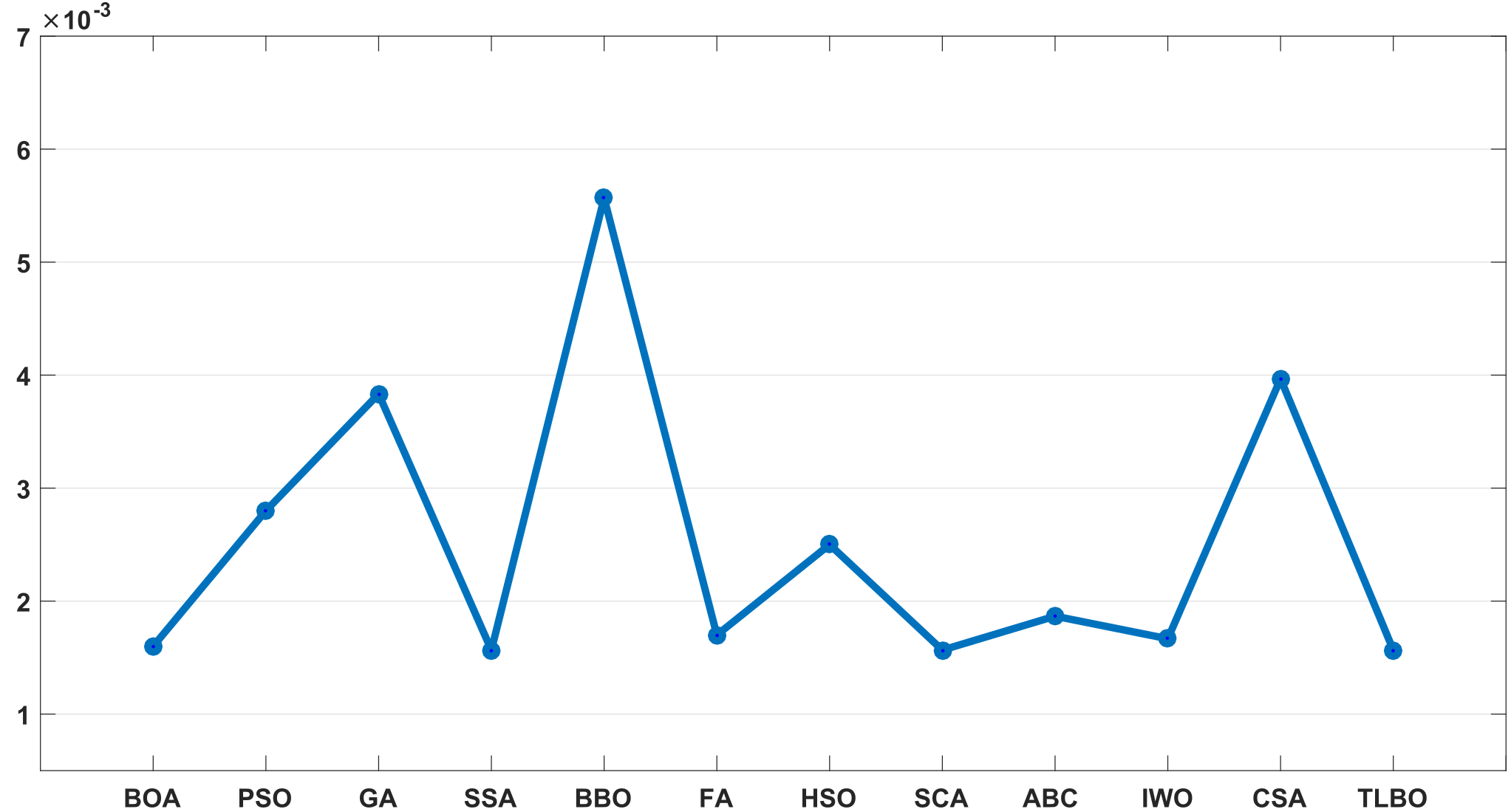}
  \centering \caption{MSEs}
     \label{mse}
\end{subfigure}
\begin{subfigure}{.4\textwidth}
\hspace*{-0.1in} 
  \includegraphics[width=1.3\linewidth, height=4.5cm]{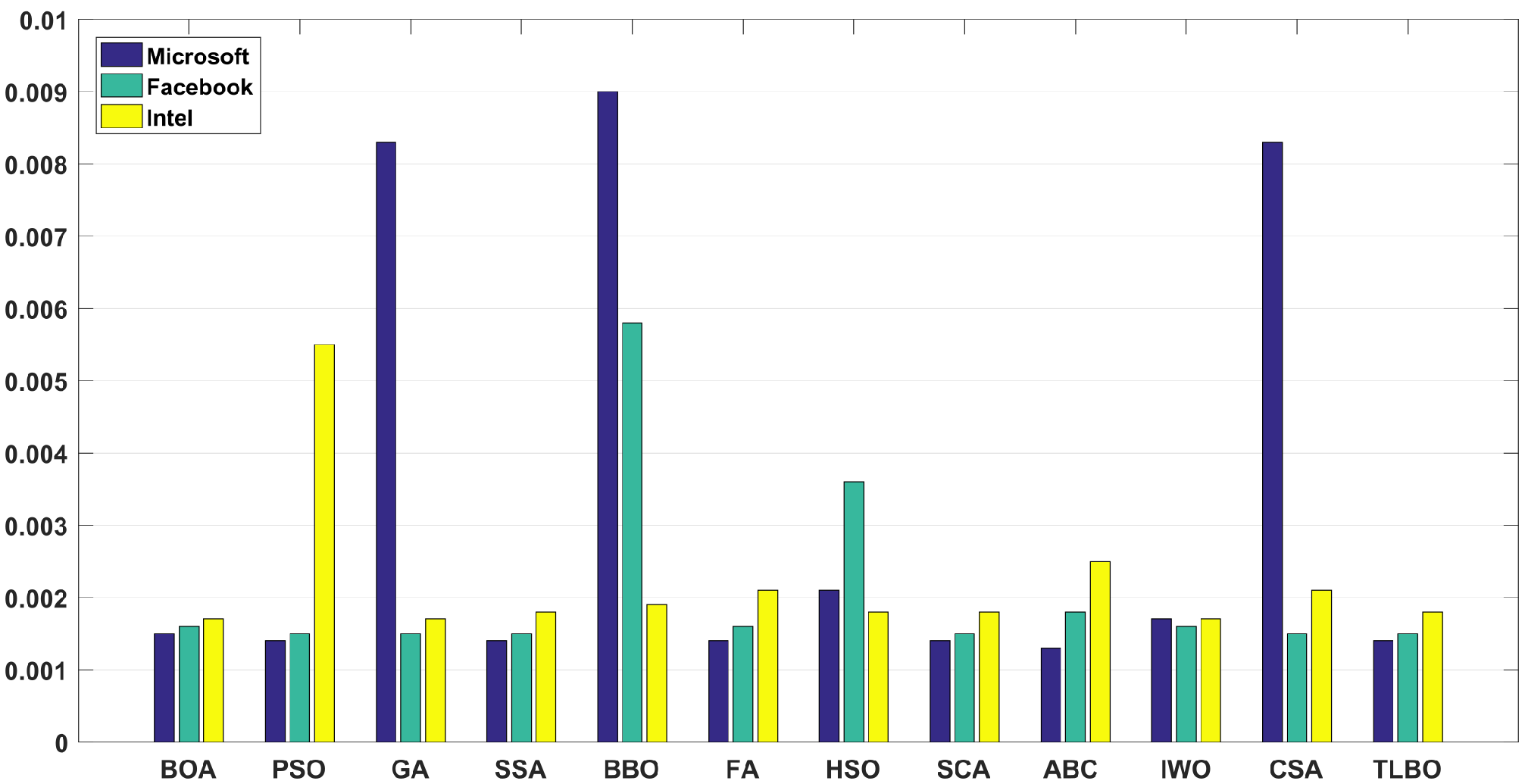}
  \centering \caption{Bar plot of MSEs}
     \label{msebar}
\end{subfigure}
\centering
\begin{subfigure}{.4\textwidth}
\hspace*{-0.9in}
  \includegraphics[width=1.3\linewidth, height=4.5cm]{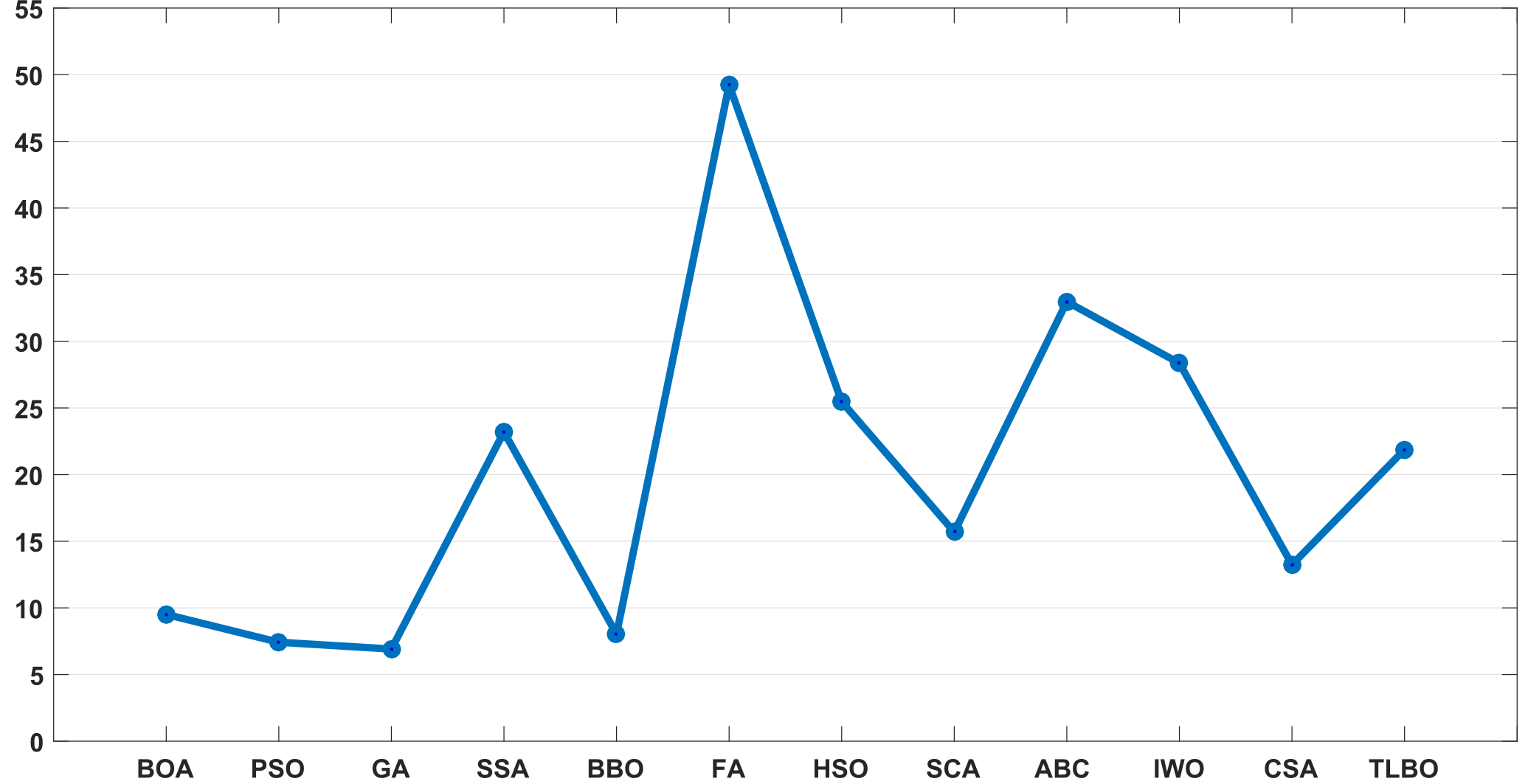}
  \centering \caption{Time Consumption}
     \label{timecosts}
\end{subfigure}
\begin{subfigure}{.4\textwidth}
\hspace*{-0.1in}
  \includegraphics[width=1.3\linewidth, height=4.5cm]{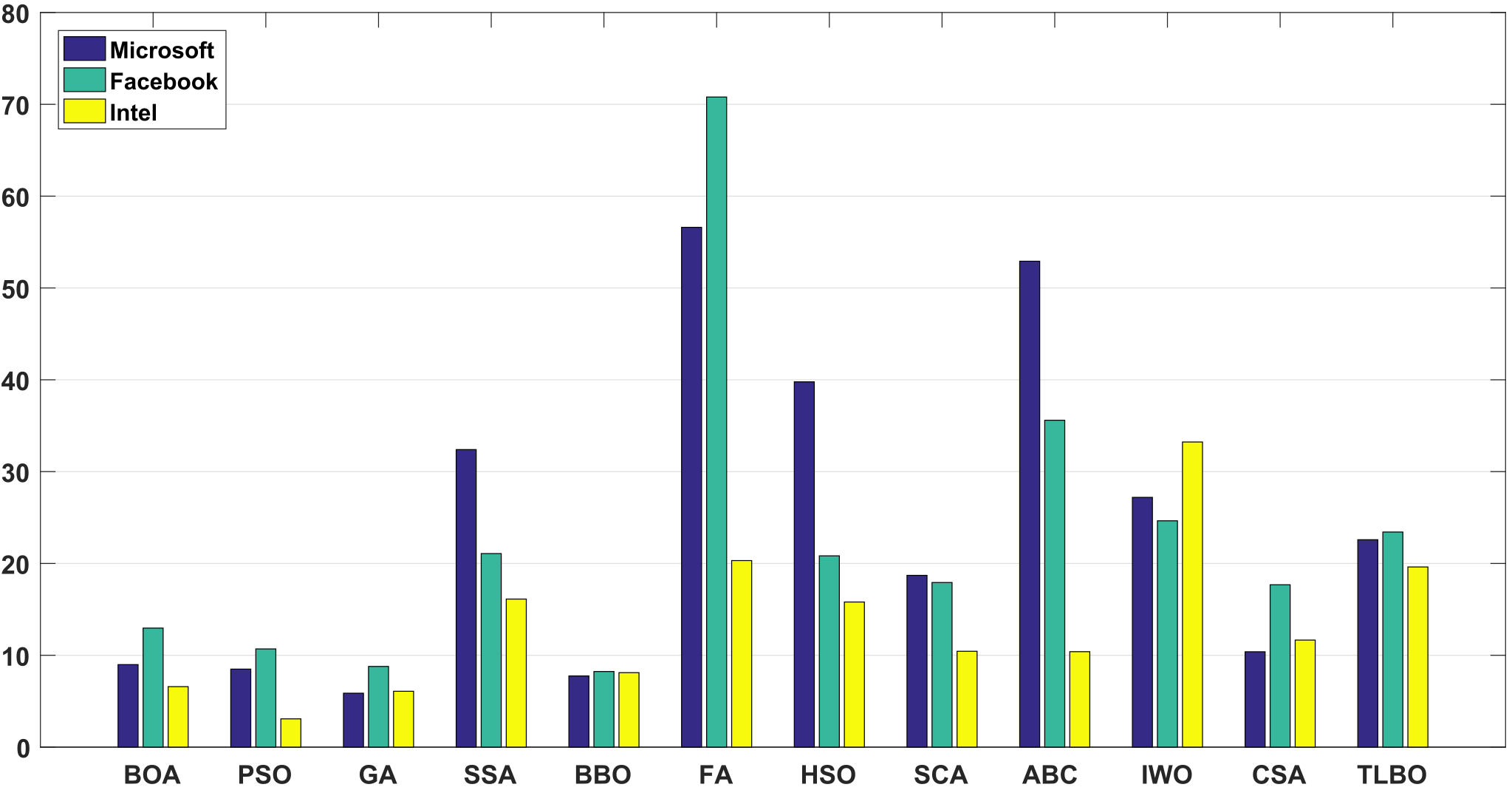}
  \centering \caption{Bar plot of Time Consumption}
     \label{timebar}
\end{subfigure}
\caption{ Fig. \ref{mape} and Fig. \ref{mapebar} compare the MAPE of the twelve SVR-based methods, Fig. \ref{mse} and Fig. \ref{msebar} compare the MSE of the SVR-based methods, and finally the average time consumption of the models is shown in Fig. \ref{timecosts} and Fig. \ref{timebar}}
\label{allaverage}
\end{figure}

Now let us investigate the accuracy measures MSE and MAPE for our model and  other eleven proposed models in more details. As it is shown in Fig. \ref{mape}, the MAPE of our proposed model achieved third rank after SSA-SVR and TLBO-SVR. We also note that even though our BOA-SVR accuracy is slightly below the other two mentioned methods, its time consumption is significantly better.
Fig.
\ref{mse}
illustrates that the BOA-SVR, SSA-SVR, SCA-SVR and TLBO-SVR algorithms achieved the best MSE accuracy compared to all the other eight methods.
Fig.
\ref{mapebar}, Fig. \ref{msebar} and Fig.  \ref{timebar} depict the bar plots of MAPE , MSE and cost time, respectively.
Based on Fig.
\ref{timecosts},
the average time consumption of our proposed BOA-SVR model ranked fourth, and competitively close to the other three methods. Although GA-SVR, PSO-SVR and BBO-SVR are computationally less expensive, but the accuracy records from Fig.
\ref{mape} and Fig. \ref{mse} show their time efficiency comes with higher accuracy costs.

Next we use the Diebold-Mariano test \cite{Diebold2002} to investigate the predictive accuracy of the SVR-based methods. Base on the Diebold-Mariano test, the null hypothesis of equality of any two given methods  
at the $5\%$ confidence level is rejected if $|DM| > 1.96$, where DM is the test statistic of the Diebold-Mariano test calculated based on the corresponding squared-error residuals.

Table \ref{DM} summarizes DM-values obtained by the Diebold-Mariano test on our three stocks. As shown in Table \ref{DM}, the forecasting accuracy of our proposed method is better than GA-SVR, BBO-SVR, HSO-SVR, IWO-SVR and CSA-SVR methods for Microsoft and there is no significant difference between BOA-SVR and the remaining models.
For Facebook, the null hypothesis of equality is rejected for BOA-SVR and HSO-SVR. In fact, BOA-SVR performed better based on forecasting accuracy. For Intel, PSO-SVR, BBO-SVR, FA-SVR, SCA-SVR and ABC-SVR have DM-test absolute value greater than 1.96, therefore there is a significant difference in terms of prediction accuracy between these models and BOA-SVR. 

To summarize, based on time efficiency, MSE and MAPE measures, we conclude that our BOA-SVR algorithm performs as one of the best models among the total set of twelve methods studied. Therefore, we believe that BOA algorithm is capable of serving as an competitively efficient parameters optimization method for Suppor Vector Regression Machines.

\input{tabs/DM.tex}

\section{Conclusion and Future Research}\label{con}

Parameter selection is considered as one of the crucial tasks in using support vector regression algorithm for time series prediction. In this study, a novel hybrid method based on SVR and butterfly optimization algorithm is presented to tune the parameters of SVR namely, penalty factor, $C$, RBF kernel function width 
parameter, $\gamma$ and radius of the epsilon tube, $\epsilon$. The proposed model has been well tested on 
three financial time series, Microsoft, Facebook and Intel, using their daily closing stock market prices. Our results are compared with eleven other metaheuristics algorithms used to estimate SVR parameters. According to the experimental results of this study, BOA-SVR is capable of tuning the parameters of the SVR model highly efficiently in terms of computational time and accuracy. This makes the BOA algorithm suitable than most of other metaheuristic algorithms for time series prediction specially in financial time series.

For future research, we see two directions to take. First, to investigate other metaheuristic optimization algorithms in estimating SVR parameters to see how they compare to the twelve methods we collected here. Second,the BOA optimization algorithm, which has been introduced very recently, can be used in other Machine Learning applications and specially for calibrating parameters of models where the estimation problem is a non-linear, non-convex problem.

\clearpage
\newpage
\appendix
\section{Appendix}

\input{tabs/mictable.tex}
\input{tabs/facetable.tex}
\input{tabs/inteltable.tex}

\newpage
\bibliographystyle{cas-model2-names}
\bibliography{cas-refs}

\end{document}

%% file: tabs/phasespace.tex
\begin{table}[width=1.0\linewidth, cols=6, pos=h]
  \centering
  \caption{Estimation of $m$ and $\tau$ for phase space reconstruction routine.}
    \begin{tabular*}{\tblwidth}{@{}CCCCCC@{} }
    \toprule
      \textbf{Parameters}     & \textbf{ Microsoft } &       & \textbf{ Facebook } &       & \textbf{ Intel} \\
    \midrule
    $m$ &  15    &       &  12    &       &  9 \\
    $\tau$ &  5     &       &  3     &       &  3 \\
    \bottomrule
    \end{tabular*}%
  \label{table 1}%
\end{table}%

%% file: tabs/DM.tex
\begin{table}[width=1.0\linewidth,cols=12]
\caption{Diebold-Mariano Test comparison of our BOA-SVR method with eleven others}
\scalebox{1.0}{
\begin{tabular*}{\tblwidth}{@{} LLLLLLLLLLLL@{} }
    \toprule
    \textbf{Stocks} & \textbf{PSO  } & \textbf{GA  } & \textbf{SSA  } & \textbf{BBO  } & \textbf{FA  } & \textbf{HSO } & \textbf{SCA } & \textbf{ABC } & \textbf{IWO } & \textbf{CSA } & \textbf{TLBO } \\
    \textbf{} & \textbf{-SVR } & \textbf{-SVR } & \textbf{-SVR } & \textbf{-SVR } & \textbf{-SVR } & \textbf{-SVR} & \textbf{-SVR} & \textbf{-SVR} & \textbf{-SVR} & \textbf{-SVR} & \textbf{-SVR} \\
    \midrule
    \textbf{Microsoft } & 1.4402 & -5.3736 & 0.9343 & -5.8809 & 1.5201 & -4.6381 & 1.2556 & 1.618 & -2.1352 & -5.7747 & 0.6382 \\
    \textbf{Facebook } & 1.3014 & 0.9342 & 1.1799 & -4.3128 & -0.0148 & -5.3942 & 0.6365 & -1.6881 & -0.1391 & 0.81  & 1.1768 \\
    \textbf{Intel } & -6.06 & 0.2778 & -1.5014 & -2.2792 & -3.3147 & -0.6705 & -2.0534 & -2.7493 & -0.6524 & -1.3309 & -1.3609 \\
    \bottomrule
    \end{tabular*}%
    }
  \label{DM}%
\end{table}%

%% file: tabs/mictable.tex
\begin{table}[width=1.0\linewidth,cols=7, pos=h]
  \centering
  \caption{Calibrated parameters as well as MSE and MAPE performance measures for Microsoft}
  \scalebox{0.90}{
\begin{tabular*}{\tblwidth}{@{} LLLLLLL@{} }    
    \toprule
    \textbf{Models} & \textbf{ C} & $\mathbf{\gamma}$ & $\mathbf{\epsilon}$ & \textbf{MSE } & \textbf{MAPE } & \textbf{Cost time} \\
    \midrule
    \textbf{BOA-SVR } & 31.2107 & 0.0273 & 0.0225 & 0.0015 & 0.0447 & 8.9851 \\
    \textbf{PSO-SVR } & 139.8311 & 0.0241 & 0.0224 & 0.0014 & 0.0445 & 8.4949 \\
    \textbf{GA-SVR } & 0.9276 & 0.4839 & 0.0163 & 0.0083 & 0.0981 & 5.8714 \\
    \textbf{SSA-SVR } & 236.0441 & 0.0043 & 0.009 & 0.0014 & 0.0427 & 32.4025 \\
    \textbf{BBO-SVR } & 0.496 & 0.5743 & 9.54E-07 & 0.009 & 0.1041 & 7.7476 \\
    \textbf{FA-SVR } & 50.5475 & 0.007 & 0.0152 & 0.0014 & 0.0422 & 56.6055 \\
    \textbf{HSO-SVR } & 3.9998 & 0.0414 & 0.0034 & 0.0021 & 0.0529 & 39.7774 \\
    \textbf{SCA-SVR} & 36.3572 & 0.0119 & 0.0156 & 0.0014 & 0.0427 & 18.7039 \\
    \textbf{ABC-SVR } & 256   & 0.0163 & 0.0251 & 0.0013 & 0.043 & 52.9133 \\
    \textbf{IWO-SVR } & 31.8184 & 0.0396 & 0.0376 & 0.0017 & 0.045 & 27.1983 \\
    \textbf{CSA-SVR} & 1.9611 & 0.4652 & 0.0019 & 0.0083 & 0.0987 & 10.3826 \\
    \textbf{TLBO-SVR } & 153.8507 & 0.0215 & 0.0223 & 0.0014 & 0.0444 & 22.5853 \\
    \bottomrule
    \end{tabular*}%
    }
  \label{mictable}%
\end{table}%

%% file: tabs/facetable.tex
\begin{table}[width=1.0\linewidth,cols=4, pos=h]
  \centering
  \caption{Calibrated parameters as well as MSE and MAPE performance measures for Facebook.}
  \scalebox{0.90}{
\begin{tabular*}{\tblwidth}{@{} LLLLLLL@{} }    
    \toprule
    \textbf{Models} & \textbf{ C} & $\mathbf{\gamma}$ & $\mathbf{\epsilon}$ & \textbf{MSE } & \textbf{MAPE } & \textbf{Cost time} \\
    \midrule
    \textbf{BOA-SVR } & 17.2259 & 0.0186 & 0.0267 & 0.0016 & 0.1284 & 12.9602 \\
    \textbf{PSO-SVR } & 155.4438 & 0.002 & 0.0191 & 0.0015 & 0.1693 & 10.6923 \\
    \textbf{GA-SVR } & 8.0186 & 0.032 & 0.0177 & 0.0015 & 0.1171 & 8.7848 \\
    \textbf{SSA-SVR } & 31.8064 & 0.0039 & 0.005 & 0.0015 & 0.123 & 21.077 \\
    \textbf{BBO-SVR } & 1.9901 & 1.597 & 9.54E-07 & 0.0058 & 0.2113 & 8.2315 \\
    \textbf{FA-SVR } & 255.0871 & 0.0016 & 0.0379 & 0.0016 & 0.1504 & 70.7953 \\
    \textbf{HSO-SVR} & 69.8398 & 0.969 & 0.032 & 0.0036 & 0.2318 & 20.8291 \\
    \textbf{SCA-SVR } & 247.8187 & 0.0016 & 0.0211 & 0.0015 & 0.1986 & 17.9262 \\
    \textbf{ABC-SVR } & 193.9844 & 0.0021 & 0.0686 & 0.0018 & 0.1241 & 35.5963 \\
    \textbf{IWO-SVR} & 214.1998 & 0.0023 & 0.0379 & 0.0016 & 0.1478 & 24.645 \\
    \textbf{CSA-SVR} & 256   & 0.0022 & 0.0239 & 0.0015 & 0.1677 & 17.68 \\
    \textbf{TLBO-SVR } & 70.454 & 0.0039 & 0.0197 & 0.0015 & 0.1631 & 23.4223 \\
    \bottomrule
    \end{tabular*}%
    }
  \label{facetable}%
\end{table}%

%% file: tabs/inteltable.tex
\begin{table}[width=1.0\linewidth,cols=4, pos=h]
  \centering
  \caption{Calibrated parameters as well as MSE and MAPE performance measures for Intel.}
  \scalebox{0.90}{
\begin{tabular*}{\tblwidth}{@{} LLLLLLL@{} }    
\toprule
    \textbf{Models} & \textbf{ C} & $\mathbf{\gamma}$ & $\mathbf{\epsilon}$ & \textbf{MSE } & \textbf{MAPE } & \textbf{Cost time} \\
    \midrule
    \textbf{BOA-SVR } & 42.799 & 0.4746 & 0.0454 & 0.0017 & 0.0451 & 6.5846 \\
    \textbf{PSO-SVR } & 95.1984 & 1.3015 & 0.1228 & 0.0055 & 0.0868 & 3.0762 \\
    \textbf{GA-SVR } & 0.3587 & 0.726 & 0.0173 & 0.0017 & 0.0456 & 6.0813 \\
    \textbf{SSA-SVR } & 204.5366 & 0.3125 & 0.0446 & 0.0018 & 0.0458 & 16.1212 \\
    \textbf{BBO-SVR } & 0.773 & 0.0568 & 9.54E-07 & 0.0019 & 0.0492 & 8.1073 \\
    \textbf{FA-SVR } & 108.5004 & 0.0066 & 0.0717 & 0.0021 & 0.0522 & 20.315 \\
    \textbf{HSO-SVR } & 9.865 & 1.0446 & 0.0462 & 0.0018 & 0.046 & 15.8023 \\
    \textbf{SCA-SVR} & 28.0421 & 0.6909 & 0.0464 & 0.0018 & 0.0459 & 10.437 \\
    \textbf{ABC-SVR } & 106.4187 & 0.9174 & 0.0509 & 0.0025 & 0.057 & 10.3934 \\
    \textbf{IWO-SVR } & 242.5888 & 0.2726 & 0.0475 & 0.0017 & 0.0451 & 33.2282 \\
    \textbf{CSA-SVR} & 2.1276 & 2.6918 & 0.0018 & 0.0021 & 0.0509 & 11.6568 \\
    \textbf{TLBO-SVR } & 86.2845 & 0.3859 & 0.048 & 0.0018 & 0.0458 & 19.6189 \\
    \bottomrule
    \end{tabular*}%
    }
  \label{inteltable}%
\end{table}%